# A Vision Based System for Monitoring the Loss of Attention in Automotive Drivers

Anirban Dasgupta, Anjith George, S L Happy and Aurobinda Routray, *Member, IEEE*

*Abstract*— On-board monitoring of the alertness level of an automotive driver has been a challenging research in transportation safety and management. In this paper, we propose a robust real-time embedded platform to monitor the loss of attention of the driver during day as well as night driving conditions. The PERcentage of eye CLOSure (PERCLOS) has been used as the indicator of the alertness level. In this approach, the face is detected using Haar-like features and tracked using a Kalman Filter. The Eyes are detected using Principal Component Analysis (PCA) during day time and the block Local Binary Pattern (LBP) features during night. Finally the eye state is classified as open or closed using Support Vector Machines (SVM). In-plane and off-plane rotations of the driver's face have been compensated using Affine and Perspective Transformation respectively. Compensation in illumination variation is carried out using Bi-Histogram Equalization (BHE). The algorithm has been cross-validated using brain signals and finally been implemented on a Single Board Computer (SBC) having Intel Atom processor, 1 GB RAM, 1.66 GHz clock, x86 architecture, Windows Embedded XP operating system. The system is found to be robust under actual driving conditions.

## I. INTRODUCTION

LOSS of attention in automotive drivers has been reported to be a major cause of road accidents [1], [2], [3], [4], [5]. Hence, to prevent such accidents, on-board monitoring of the alertness level in automotive drivers is necessary. Alertness level can be assessed using different measures [6] such as Electroencephalogram (EEG) signals [7], ocular features [8], blood samples [8], speech [9], [10] and others. The EEG based method has been reported to be highly authentic for estimating the state of drowsiness [7]. However, on-board measurement of EEG signals may cause discomfort to the driver and also cause artefacts. In [11] Qiang Ji reported a review of non-invasive techniques to monitor human fatigue. The report states that the potential non-invasive fatigue monitoring measures are eyelid movements, eye gaze, head movement, and facial expressions. The vision-based approaches, being non-contact in nature, appear to be promising for estimating the aforementioned parameters. There has been some studies [5], [12], [13], [14] with regard to eyelid movements such as blink frequency, Average Eye-Closure Speed (AECS), PERCLOS, etc. as quantitative measures of the drowsiness level of an individual. Of them, PERCLOS is reported to be the best and most robust measure for fatigue detection [15], [16]. PERCLOS as a drowsiness metric was established in [12] by Wierwille *et al.* The authenticity of PERCLOS as an effective indicator of drowsiness level has been validated in [17] [18]. PERCLOS can be estimated from continuous video sequences of the eye images [19]. PERCLOS [20] may be defined as the proportion of time the eyelids are at least 80% closed over the pupil. It is a temporal ratio over a given time window. There are several steps involved before estimating the accurate PERCLOS value. The first step is face detection followed by eye detection and eye state classification. A Region Of Interest (ROI) is selected on the detected face boundary to locate the eyes and then classify the eye state as open or closed. In the present case, closed eye implies that the eyelid is closed at least 80%. Finally, the PERCLOS value is calculated as follows.

$$P = \frac{E_c}{E_t} \qquad (1)$$

In (1), $P$ is the PERCLOS value, $E_c$ is the number of closed eye count over a predefined interval while $E_t$ is the total eye count in the same given interval. Literature [20] states that a higher value of $P$ indicates higher drowsiness level and vice versa.

In this paper, we have designed real-time algorithms to estimate the PERCLOS of a driver taking into the considerations certain issues prevailing in the literature and implemented them on a real-time embedded platform suitable for automotive environment. The accuracy of our algorithm in estimating the level of drowsiness has been validated by EEG signals in our previous works [21], [22].

This paper is organized as follows. Section II discusses a review of earlier works on eye detection and PERCLOS estimation. Section III presents the real-time face detection

This work is supported in part by the Department of Electronics and Information Technology, Ministry of Communications and Information Technology, Government of India.

Anirban Dasgupta is an MS student in the Electrical Engineering Department, Indian Institute of Technology, Kharagpur, India (e-mail: anirban1828@gmail.com).

Anjith George is a Research Scholar in the Electrical Engineering Department, Indian Institute of Technology, Kharagpur, India (e-mail: anjith2006@gmail.com).

S. L. Happy is an MS student in the Electrical Engineering Department, Indian Institute of Technology, Kharagpur, India (e-mail: happpyhoppy@gmail.com).

Aurobinda Routray is working as Professor in the Electrical Engineering Department, Indian Institute of Technology, Kharagpur, India (e-mail: aroutray@iitkgp.ac.in).

algorithm for estimating PERCLOS, addressing the issues of face rotation. In Section 0, we discuss a method to account for varying illumination levels. Section 0 discusses the use of Kalman filter for face tracking. In Section VI, we discuss real-time eye detection for both day and night driving conditions. In Section VII, eye state classification is discussed using Support Vector Machine (SVM). Section VIII discusses the on-board testing during day and night driving conditions. Finally, Section IX concludes the paper.

## II. LITERATURE REVIEW

Several approaches have been reported to develop systems for on-board monitoring of the alertness level in automotive drivers. The work in [23] reports a system named as Driver AssIsting SYstem (DAISY) as a monitoring and warning aid for the driver in longitudinal and lateral control on German motorways. The warnings are generated based on the knowledge of the behavioural state and condition of the driver. However, this is a model based approach and proper on-board testing and validation has not been discussed. Singh et al. [24] proposed a non-invasive vision-based system for the detection of fatigue level in drivers. The system uses a camera that points directly towards the driver's face and monitors the driver's eyes in order to detect micro-sleeps. The system is reported to detect the eyes at 10 fps and track at 15 fps. The accuracy of the system is about 95% with small movement of head bearing a tolerance on head tilt up to 30 degrees. However in practical driving scenario, this system has drawbacks where driver's head movement introduces inefficiency of the proposed system. Robotics Institute in Carnegie Mellon University has developed a drowsy driver monitor named Copilot [25], which is a video-based system and estimates PERCLOS. A similar system called Driver Assistance System (DAS) has been developed by a group at the Australian National University [26]. It uses a dashboard-mounted head-and-eye-tracking system to monitor the driver. The Distillation algorithm has been used to monitor the driver's performance. Feedback on deviation in lane tracking is provided to the driver using force feedback to the steering wheel which is proportional to the amount of lateral offset estimation by the lane tracker. However in the systems [23], [24], [25] the issues such as illumination variation and head rotations have not been adequately addressed. In [27], Ji et al. describe a real-time online prototype driver-fatigue monitoring system using remotely located charge-coupled-device cameras equipped with active infrared illuminators. They have combined eyelid movement, gaze movement, head movement, and facial expression for assessment of the level of alertness of the driver. They have compared the drowsiness detection using AECS and PERCLOS. Smith et al. [5] describe a system in which the driver's head is detected using colour predicates and tracked using optical flow. The system described is reported to be robust against face rotations, as well as head and eye occlusions. Eye occlusions are detected by the likelihood of rotational occlusion of eyes. The algorithm gives an approximate 3D gaze direction with a single camera. The size of head is assumed constant among different persons so the gaze direction and can be estimated without prior knowledge of the distance between head and camera. However the accuracy of the system is illumination dependent. Zhu and Ji [13] present a real-time system for on-board monitoring the driver using active infrared illumination. Visual cues such as PERCLOS, AECS, 3D head orientation and facial expressions are fused with a Bayesian framework to obtain the estimation of fatigue level. In [28], Hanowski et al. have carried out investigations to detect the drowsiness in long haul truck drivers. They recorded the vehicle performance parameters such as velocity, lateral and longitudinal acceleration, steering position, and brake pedal activation using appropriate sensors. The trucks had cameras that provided the exterior views of the driving environment as well as the images of the driver's face. They finally analyzed the driving behavior and human-factors issues for truck driving. Bergasa et al. [29] have developed a system which monitors the driver's fatigue level with six different parameters - PERCLOS, eye closure duration, blink frequency, head nodding frequency, face position, and fixed gaze. These different cues are combined together using a fuzzy classifier to get the alertness level of the driver. Bright-pupil dark-pupil method is used for the detection pupil. The system is based on active illumination and hence requires special hardware support. In [30], Eriksson and Papanikolopoulos propose a system to locate and track the eyes of the driver. They used a symmetry-based approach to locate the face for subsequent detection and tracking of eyes. Template matching has been used for eye state classification. Matthew Sacco et al. [31] have presented a real-time system for classification of alertness of drivers. Viola-Jones frame work has been used for the detection of eyes, and mouth, followed by tracking of these features using Template Matching. The final decision about fatigue level is taken by considering three parameters - PERCLOS, average eye closure interval and the degree of mouth opening. SVM is used to fuse these three different features. However the performance with head rotations and effect of variations in illumination including darkness during night driving conditions has not been reported in this work.

From the above review, PERCLOS has been selected as the final indicator of the level of drowsiness. It is evident that the eye region has to be classified as open or closed to estimate PERCLOS which require accurate eye detection. Several approaches have been reported to detect eyes and compute PERCLOS. In [32], Huang et al. used optimal wavelet packets for representing eyes, thereby classifying the face area into eye and non-eye regions using Radial Basis Functions. The optimal wavelet packets are selected by minimizing their entropy. They have obtained an eye detection rate of 82.5% for eyes and 80% for non-eyes using 16 wavelet coefficients whereas 85% for eyes and 100% for non-eyes using 64 wavelet coefficients. However in this paper the real-time performance of the algorithm has not been established. In [33], Feng et al. used three cues for eye detection using gray intensity images. These are face intensity, the estimated direction of the line joining the centers of the eyes and the response of convolving the proposed eye variance with the face image. Based on the cues, a cross-validation process has been carried out to generate a list of possible eye window pairs. For each possible case, the variance projection function is used for eye detection and verification. They obtained a detection accuracy of 92.5% on a face database from MIT AI laboratory, which contains 930 face images with different orientations and hairstyles captured from different persons. However, low accuracy in eye detection was obtained in situations such as rotation of the face or occlusion of eyes due to eyebrows. This also does not address the real time implementation issues. In [34], Sirohey *et al.* used two methods of eye detection in facial images. The first is a linear method using filters based on Gabor wavelets. They obtained a detection rate 80% on Aberdeen dataset and 95% on the dataset prepared by them. The second was a nonlinear filtering method to detect the corners of the eyes using color-based wedge shaped filters, which has a detection rate of 90% on the Aberdeen database. The nonlinear method performed better over the linear method in terms of false alarm rate. However, the methods discussed are yet to be tested for real-time applications. In [35], Viola *et al.* described a robust real-time object detection framework based on Haar-like features. They introduced integral image representation for fast computation of these features. Then, they used the AdaBoost algorithm [36] to select a small number of features out of many. Finally, they combined each weak classifier to form a cascade, thereby making it very robust against variations. They obtained a detection rate of 93.7% on the MIT+CMU face database. The advantage of using Haar-like features is that apart from being a real-time algorithm, it is also scale invariant. Zhou *et al.* [37] defined the Generalized Projection Function (GPF) as a weighted combination of the Integral Projection Function (IPF) and the Variance Projection Function (VPF). They proved that IPF and VPF are special cases of GPF when the parameter $\alpha$ is 0 and 1, respectively. They also got hybrid projection function (HPF) from GPF using $\alpha$ as 0.6. From HPF, they found that eye area is darker than its neighboring areas and the intensity of the eye area rapidly changes. They achieved best detection rates with $\alpha = 0.6$, on BioID, JAFFE and NJU face databases with a detection rate of 94.81%, 97.18% and 95.82% respectively. In [38], Song *et al.* presented a new eye detection method which consists of extraction of Binary Edge Images (BEIs) from the grayscale face image using multi-resolution wavelet transform. They achieved an eye detection rate of 98.7% on 150 Bern images with variations in views and gaze directions and 96.6% was achieved on 564 images in AR face databases with different facial expressions and lighting conditions. However the algorithm has not been tested in real-time and is prone to illumination variation as it uses intensity information for eye localization. The illumination variation issues have not been properly addressed in this work. In [39], Wang *et al.* studied the impact of eye locations on face recognition accuracy thereby introducing an automatic technique for eye detection. They used Principal Component Analysis (PCA) and PCA combined with Linear Discriminant Analysis (LDA) technique for face recognition. Subsequently, AdaBoost was used to learn the discriminating features to detect eyes. They tested their technique on FRGC 1.0 database and achieved an overall accuracy of 94.5% in eye detection. They also found that only PCA recognizes faces better than PCA combined with LDA. However, their technique has been tested for offline applications. In [40], Asteriadis *et al.* localized eyes in the face based only on geometrical information. They obtained about 99.5% eye localization accuracy on BioID database and 99.2% accuracy on XM2VTS database.

In [41], Eriksson *et al.* used template matching for eye tracking to detect fatigue in human drivers. They used eye blink rate to detect the fatigue level. However, eye blink can vary from person to person and is not an accurate indicator of fatigue. Orazio *et al.* [42] used a neural classifier to detect eyes for assessing driver's alertness. The classifier contained 291 eye images and 452 non-eye images as training. Their algorithm is found to work even on faces having glasses with variations in illumination. The algorithms are tested offline. The issues like vibration of the vehicle are not considered in the algorithm. Hong *et al.* [43] developed an embedded system to detect driver's drowsiness by computing the PERCLOS. They detected face using Haar-like features and localized the eye using horizontal and vertical variance projection. They used Camshift algorithm to track the eyes, thereby limiting the search area by Camshift prediction in YCbCr plane. They classified the eye state by thresholding eye region, then applying Laplacian on the binary image and finally using a complexity function. They obtained an accuracy of about 85%. The algorithm has been tested for real-time applications. However, being dependent on color information, the algorithm is illumination dependent. Its performance may be poor for on-board situations where occurrence of shadows will cause problems. Liying *et al.* [44] detected the eyes, based on skin color segment and identified the eye's condition by combining the color segmentation and morphological operations. However, the algorithm being dependent on color information will also be prone to variations in lighting conditions. Face rotation is also not considered in the work. . Hansen et al. [45] developed an improved likelihood for eye tracking in uncontrolled environments. They have used BPDP method to detect the eyes.

From the above cited literature review, the issues pertaining to the real time detection of eyes and subsequent estimation of PERCLOS values for real time and in-vehicle applications are as follows:

- Improving the frame rate within real-time constraints
- Detection of face and eyes with variation in illumination levels
- Detection with different face orientations
- Detection in presence of noise and interference due to vibration
- Detection of eyes with spectacles

In the present work, some of the above issues have been addressed and a real-time algorithm has been developed and implemented on an embedded platform for estimating the PERCLOS value. Haar-like features have been used to detect the face. The potential region of finding the face is tracked using a Kalman Filter thereby reducing the search space for face detection and increasing the real-time performance. Affine and Perspective transformations have been used to consider the in-plane and off-plane rotations of the face respectively. BHE is used to compensate for the variation in lighting conditions. Eye detection is carried out using PCA during daytime and LBP based features during nighttime. The eye state is classified as open or closed using SVM. A 66.67% overlapping window of 3 minutes duration [46] is selected to compute PERCLOS. Based on a threshold of 15%, the driver's alertness state is classified as drowsy or alert. A voice alarm is set to warn the driver, in the event of PERCLOS exceeding the threshold. Testing has been carried out in laboratory as well as on board for both day as well as night driving conditions. The limitation of the algorithm is that cases of drivers wearing spectacles have not been considered. However, preliminary work has been reported in [47] for detection of spectacles.

## III. REAL-TIME FACE DETECTION

The first approach in the estimation of PERCLOS is face detection. This section describes the algorithm for face detection under varying conditions.

### A. Face detection using Haar-like features

Haar-like features [35] are rectangular features in a digital image which are used in object detection. Viola *et al.* [35] combined such features in a cascade of classifiers to detect objects in real-time. Fig. 1 shows some Haar-like features. The value of a rectangular Haar-like feature is the difference between the sum of the pixels within the white rectangular regions and that within the black rectangular regions.

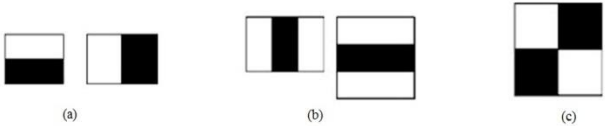

**Fig. 1 (a) two rectangle features (b) three rectangle features (c) four rectangle features**

For fast computation of rectangular Haar-like features, integral images [35] are computed. Each element of the integral image contains the sum of all pixels located on the up-left region of the image. Using this, we compute the sum of rectangular areas in the image, at any position or scale. The integral image $I(x, y)$ is computed efficiently in a single pass over the image using:

$$I(x,y) = i(x,y) + I(x-1,y) + I(x,y-1) - I(x-1,y-1) \quad (2)$$

In (2), $I(x,y) = \sum_{\substack{x' \le x \\ y' \le y}} i(x', y')$ and $i(x, y)$ is the intensity at point $(x, y)$. There are huge number of rectangle features associated with each image sub-window [35], which is far larger than the number of pixels. Even though each feature can be computed very efficiently, computing the complete set is cumbersome. Viola *et al.* [35] reported that a very small number of these features can be combined to form an effective classifier. They have used Adaptive Boosting (AdaBoost) for the purpose of selecting such features. After obtaining a weak classifier, several weak classifiers are cascaded to obtain a strong classifier to reject as many false negatives as possible at the earliest. Training of Haar-like feature based classifiers includes several steps as described in [48]. The optimal sets of training parameters are selected by the method used in our earlier work [49].

Once the classifier is tested offline, it is deployed to detect the face and eyes from images taken from the camera in real time. The processing speed was found to be 3 fps on an SBC having 1GB RAM and Intel Atom Processor with 1.6 GHz clock. Such a speed is low for accurately estimating the PERCLOS value. Hence, there was a need to improve the real-time performance of the algorithm without compromising on accuracy. It is apparent that once the integral image has been calculated, any one of these Haar-like features can be calculated at any scale or location in constant time [35]. This means, to speed up the detection, the time taken for the computation of the integral image has to be minimized. This is achieved by down sampling the incoming frame. Table I states the algorithm of face detection for estimating the PERCLOS.

TABLE I
ALGORITHM FOR FACE DETECTION

| |
|---|
| 1) The original image from the camera is saved. The image is scaled down by a factor of $k$. |
| 2) Face is detected in the down sampled image using the Haar classifier. |
| 3) An ROI is selected as the upper half of the detected facial image, based on morphology of human face. |
| 4) ROI co-ordinates obtained in the down sampled image are remapped on to the original frame as shown in Fig. 2 |
| 5) Eyes are then detected in the selected ROI, using methods as described later in this paper. |

This method achieves a higher detection rate, which is dependent on its Scale Factor (SF), primarily because of the faster processing at a lower resolution. Bicubic interpolation [50] is used for down sampling the frames. Fig. 2 shows our scheme of down sampling, extraction and remapping of the ROI. In this figure, $a'_i s$ & $b'_i s$ for $i = 1,2,3,4$ are obtained from the face detector based on Haar Classifier. ROI co-ordinates $c_2$ & $c_3$ are obtained as $c_2 = \frac{b_1+b_2}{2}$ and $c_3 = \frac{b_3+b_4}{2}$ respectively. The ROI co-ordinates $d'_i s$ & $e'_i s$ are obtained using the following relations as given (3) and (4).

If $D = \begin{bmatrix} d_1 \\ d_2 \\ d_3 \\ d_4 \end{bmatrix}, E = \begin{bmatrix} e_1 \\ e_2 \\ e_3 \\ e_4 \end{bmatrix}, A = \begin{bmatrix} a_1 \\ a_2 \\ a_3 \\ a_4 \end{bmatrix}$ & $B = \begin{bmatrix} b_1 \\ c_2 \\ c_3 \\ b_4 \end{bmatrix}$, then

$$D = k \times A \quad (3)$$
$$E = k \times B \quad (4)$$

As the SF increases, as expected the frame rate is found to increase. However, the fall in accuracy puts a limit to the increasing value of SF. Hence an optimal SF has to be chosen as a trade-off between speed and accuracy. In the following section, we describe how such an SF is found out experimentally.

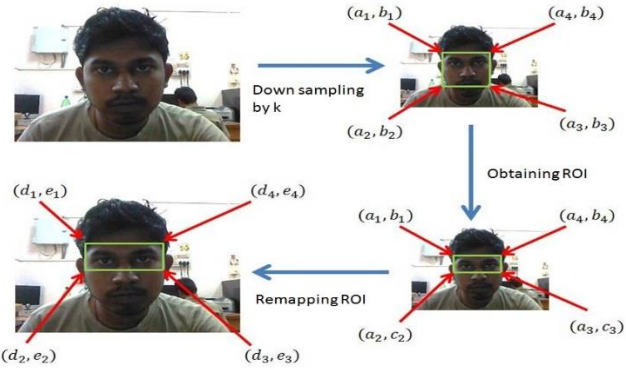

**Fig. 2 Scheme of eye detection using down sampling, extraction and remapping of ROI**

*B. Effect of SF on runtime*

To study the variation in speed of execution with the number of pixels, an experiment is conducted with six subjects and the face detection algorithm is executed. Videos of facial and non-facial images of six subjects are recorded under laboratory conditions at 30 fps with resolution of $640 \times 480$ pixels. The processing is carried out offline on a computer with Intel dual core processor, clock 2.00 GHz, 2GB RAM. The incoming frame is down sampled with SF's of 2, 4, 6, 8 and 10. The video processing speed for each SF is noted down. Fig. 3 shows some sample images extracted from these videos.

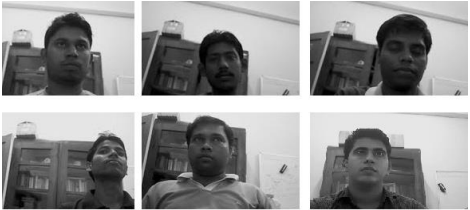

**Fig. 3 Sample images used for selecting optimum SF for face detection**

The average execution speed of face-detection in all the subjects for each SF is plotted as in Fig. 4. It reveals that the speed increases almost linearly with increasing SF. However, SF cannot be gone on increasing as accuracy puts a restriction on the SF in the next section.

*C. Effect of SF on accuracy*

The accuracy of detection is analyzed by plotting Receiver Operating Characteristics (ROC) curves and then calculating the Area Under the Curve (AUC). The ROC curve, [51] is the plot of true positive rate (tpr) vs. false positive rate (fpr) and is required to study the relative accuracies for each SF.

The videos used for run-time analysis are also used for the analysis of accuracy vs. SF. First the frames are extracted from the recorded video. The images are manually marked to take into account the presence of face to save the ground truth.

The face detection results are compared with the ground truth to obtain the number of true positives (tp), false positives (fp), true negatives (tn) and false negatives (fn). The *tpr* and *fpr* are calculated. Fig. 5 shows the ROC comparisons for each SF. The AUC versus SF curve is shown in Fig. 6.

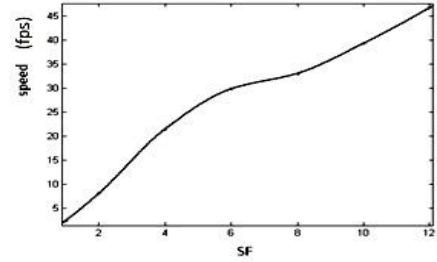

**Fig. 4 SF vs. processing speed**

It can be seen from Fig. 5 and Fig. 6 that the accuracy in face detection remains almost constant up to an SF of 6 and then decreases with a further increase in SF. Hence, an SF of 6 is taken as the optimal trade-off between run-time and accuracy.

However, the face detection algorithm, considered so far shows robustness against frontal faces only. But in practical driving situations, the driver's face may have rotations. Two different transformations have been used to compensate the effect of rotation of the driver's face.

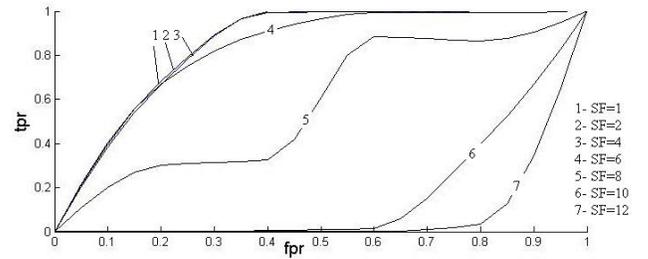

**Fig. 5 ROC comparisons for each SF**

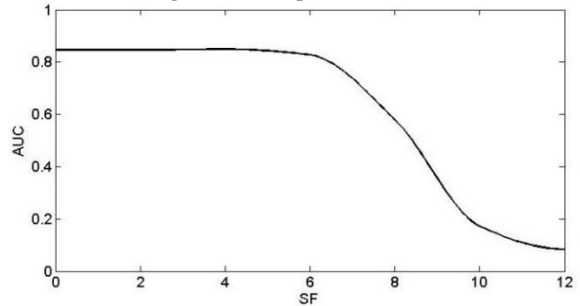

**Fig. 6 SF vs. AUC**

*A. Affine Transformation Based Detection of In-plane Rotated Faces*

In [52], an affine transformation has been used for the detection of in-plane rotated faces. The rotation matrix $R$ is found from the angle of rotation $\theta$ of the image matrix as shown in (5). If a point on the input image is $A = \begin{bmatrix} x \\ y \end{bmatrix}$ and the corresponding point on the affine transformed image is $B = \begin{bmatrix} x' \\ y' \end{bmatrix}$, then $B$ can be related to $A$ using (6).

$$R = \begin{bmatrix} \cos\theta & -\sin\theta \\ \sin\theta & \cos\theta \end{bmatrix} \quad (5)$$
$$A = RB \quad (6)$$

The angles of rotation $\theta$ are assigned values of $\pm 30^o$ successively. After the face detection, it returns the ROI for eye detection in a de-rotated form.

Fig. 7 shows in-plane rotated face detection using an affine transformation of the input image. The algorithm for detection of in-plane rotated faces is shown in Table II.

The affine transformation only occurs in the event the frontal face is not detected by the Haar classifier and hence the average processing speed is not affected much. In the following section, perspective transformation based detection of off-plane rotated faces is discussed.

TABLE II
ALGORITHM: DETECTION OF IN-PLANE ROTATED FACES

1) The Haar classifier checks for the presence of frontal faces in the input image.
2) If frontal face is found it follows the scheme as given in Table 3.
3) If frontal face is not found, we form a rotational matrix **R** with $\boldsymbol{\theta} = \pm 30^o$, to transform the image using (37). The angle of $\boldsymbol{\theta} = \pm 30^o$ is chosen as because the Haar classifier in this case has been developed with inherent rotational features up to $\pm 30^o$.
4) The Haar classifier searches for the face in the transformed image.
5) If for any of the set values of $\boldsymbol{\theta}$, the Haar classifier does not detect the face, the algorithm concludes that no face is found.

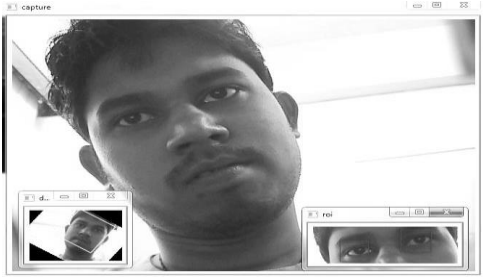

**Fig. 7 Affine Transformation Based Detection of In-plane Rotated Faces**

*B. Perspective Transformation Based Detection of Off-plane Rotated Faces*

Perspective transformation is robust and computationally inexpensive method for aligning the head for small off-plane rotations [53]. It is a mapping of one view of a face to another view. The off-plane rotated image $I(x', y')$ can be related to its original frontal image $I(x, y)$ using

$$x' = \frac{a_0 x + a_1 y + a_2}{a_3 x + a_4 y + 1} \quad (9)$$

$$y' = \frac{a_5 x + a_6 y + a_7}{a_3 x + a_4 y + 1} \quad (10)$$

The coefficients $a_0 - a_7$ are estimated using non-linear least squares method as in [53]. On obtaining the coefficients, the desired perspective mapping is obtained.

The driver's face can have two such off-plane rotations as shown in Fig. 8. Using the perspective transformation, such rotations can be compensated for small off-plane rotations. However, during normal driving, cases of off-plane rotations are observed very rarely and are restricted within $\pm 15^o$.

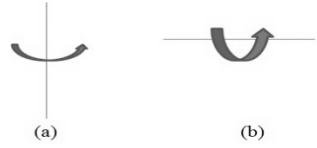

**Fig. 8 Possible Off plane rotations of driver's face**

The other issue is variation in illumination which poses serious problems to the detection accuracy.

IV. COMPENSATION OF EFFECT OF LIGHTING CONDITIONS

The detection accuracy is found to be low when the lighting conditions are extremely bright or dark. Bi-Histogram Equalization (BHE) [54] of the pixel intensities has been used to compensate for the varying illumination levels. In case the face is not detected even after the affine or perspective transformations, the algorithm considers the problem as an illumination variation issue and performs BHE as a preprocessing step. In this method, the input image is decomposed into two sub-images. One of the sub-images is the set of samples less than or equal to the mean image intensity whereas the other one is the set of samples greater than the mean image. The BHE is used to equalize the sub-images independently based on their respective histograms under the constraint that the pixel intensities in the first subset are mapped onto the range from the minimum gray level to the input mean intensity while the pixel intensities in the second subset are mapped onto the range from the mean intensity to the maximum gray level. Hence, the resulting equalized sub-images are bounded by each other around the input mean. Hence the mean intensity is preserved which makes the method illumination invariant.

The input image $X$, is subdivided into two sub-images
$$X_L = \{X(i,j) | X(i,j) \leq X_m, \forall X(i,j) \in X\} \quad (7)$$

$$X_U = \{X(i,j) | X(i,j) > X_m, \forall X(i,j) \in X\} \quad (8)$$

$X_L$ and $X_U$ based on the mean intensity image $X_m$ such that $X = X_L \cup X_U$, where

Now, histogram equalization is carried out on images $X_L$ and $X_U$ separately. A data set of 400 images is prepared of on-board images of a driver under extremely varying lighting conditions. Some images are shown in

Fig. **9**. The dataset is tested with our algorithm both before and after applying BHE. Using BHE, the detection accuracy was found to improve from 92% to 94%. Fig. 10 shows the comparison of normal and equalized images of a driver under on-board conditions. However, the results reveal that there is scope of further improvement to compensate varying illumination levels.

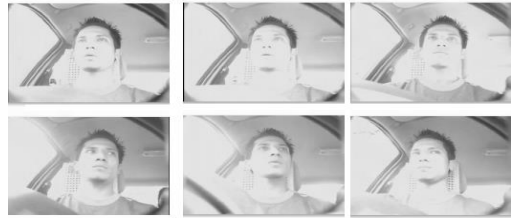

**Fig. 9 Sample images taken on-board under bright light**

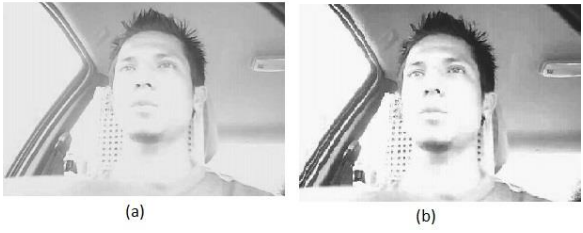
**Fig. 10 (a) Original and (b) Bi-Histogram Equalized images**

The pre-processing steps such as the geometric transformations and BHE, improves the accuracy but at the same time reduces the frame rate, owing to their computational burden. Hence, Kalman Filter based face tracking is used to reduce redundant search space by using temporal information. The following section discusses the tracking of facial region to improve the speed of processing further.

## V. KALMAN FILTER BASED FACE POSITION ESTIMATION.

In the context of automotive drivers, the face position does not change abruptly in consecutive frames. Based on this information, the probable face region can be estimated and hence the search region of the Haar classifier may be reduced. As the position and size of the face is known from the detection results of the Haar classifier, a rectangle is circumscribed around the detected face for subsequent face detection. In this paper, the circumscribed rectangle is termed as Region Of Tracking (ROT) and is twice the size of the detected face.

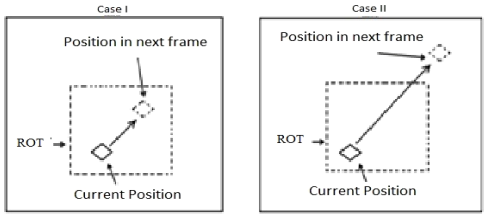
**Fig. 11 Tracking of object in continuous frames**

Fig. 11 shows two cases in tracking. Case I predicts the face within the ROT whereas in Case II, the position of the face is predicted in a region out of the ROT owing to the fast motion of face. Under the assumption that the driver's face position does not change abruptly in two consecutive frames with a proper prediction technique, Case II shall be avoided. There are several methods available for object tracking like mean shift tracking [55], optical tracking [56] and Kalman filter based tracking [57]. The first two methods rely on image intensity values in subsequent frames and hence is illumination dependent thereby making them unsuitable for our application. In this work, Kalman filter based method has been selected.

### A. Defining the Model
The system model is given below:
$$x_{k+1} = Fx_k + w_k \quad (11)$$
In (11), $x_k$ is the state vector at the $k^{th}$ frame. $F$ is the state transition matrix. The process noise $w_k$ is assumed to be Additive White Gaussian Noise (AWGN) with zero mean and covariance as:
$$E\{w_k w_k^T\} = Q \quad (12)$$

In (12), $E\{\}$ is the expectation operator. The measurement equation is stated as:
$$z_{k+1} = Hx_{k+1} + v_{k+1} \quad (13)$$
Here, $z_k$ is the measurement vector at the $k^{th}$ frame. $H$ is the observation matrix. The measurement noise $v_k$ is also assumed to be AWGN with zero mean and covariance as given below.
$$E[v_k v_k^T] = R \quad (14)$$
After defining the model, we now define the system to track the face using Kalman Filter.

### B. Defining the System
The matrices $F, H, Q$ and $R$ and suitable initial value for $P_k$ is defined. The measurement vector is defined as below:
$$z_k = [x_{1,k} \ y_{1,k} \ x_{2,k} \ y_{2,k}]^T \quad (15)$$

where $(x_{1,k}, y_{1,k})$ and $(x_{2,k}, y_{2,k})$ are the co ordinates of the left lower and right top corner of the face bounding rectangle obtained from Haar classifier based face detector at the $k^{th}$ frame.
The state vector is selected as shown below
$$x_k = [x_{1,k} \ y_{1,k} \ x_{2,k} \ y_{2,k} \ u_{x,k} \ u_{y,k}]^T \quad (16)$$

Here, $u_{x,k}$ and $u_{y,k}$ are velocities of the centre of the box in the horizontal and vertical directions respectively as defined as:
$$u_{x,k} = \frac{x_{1,k} + x_{2,k}}{2} - \frac{x_{1,k-1} + x_{2,k-1}}{2} \quad (17)$$
$$u_{y,k} = \frac{y_{1,k} + y_{2,k}}{2} - \frac{y_{1,k-1} + y_{2,k-1}}{2} \quad (18)$$
The process noise matrix $Q$ measures the variability of the input signal from the ideal transitions defined by the transition matrix. The process noise matrix $Q$ is defined as given below.
$$Q = \begin{bmatrix} 1/4 & 0 & 0 & 0 & 1/2 & 0 \\ 0 & 1/4 & 0 & 0 & 0 & 1/2 \\ 0 & 0 & 1/4 & 0 & 1/2 & 0 \\ 0 & 0 & 0 & 1/4 & 0 & 1/2 \\ 1/2 & 0 & 1/2 & 0 & 1/4 & 0 \\ 0 & 1/2 & 0 & 1/2 & 0 & 1/4 \end{bmatrix} \times 10^{-2} \quad (19)$$
where time step, $k = 1$ and average acceleration $a$ in both $x$ and $y$ directions estimated from pre-recorded video sequences as, $a = 0.1$. The measurement matrix (13) $H$ is given as:
$$H = \begin{bmatrix} 1 & 0 & 0 & 0 & 0 & 0 \\ 0 & 1 & 0 & 0 & 0 & 0 \\ 0 & 0 & 1 & 0 & 0 & 0 \\ 0 & 0 & 0 & 1 & 0 & 0 \end{bmatrix} \quad (20)$$
The matrix $R$ defines the error in measurement. In this case, the measurement is the face location from Haar based face detector whose accuracy is found to be around 95% [35]. Assuming the error is Gaussian distributed, a variance of 6.5 pixels is obtained for each of the coordinates. The error covariance is given as.

$$R = \begin{bmatrix} 1 & 0 & 0 & 0 \\ 0 & 1 & 0 & 0 \\ 0 & 0 & 1 & 0 \\ 0 & 0 & 0 & 1 \end{bmatrix} \times 42.25 \quad (21)$$

The estimated covariance matrix $P$ is a measure of the accuracy of $x_k$ at the $k^{th}$ frame. It is adjusted over time by the filter $P$ is initialized as:

$$P = \begin{bmatrix} 1 & 0 & 0 & 0 & 0 & 0 \\ 0 & 1 & 0 & 0 & 0 & 0 \\ 0 & 0 & 1 & 0 & 0 & 0 \\ 0 & 0 & 0 & 1 & 0 & 0 \\ 0 & 0 & 0 & 0 & 1 & 0 \\ 0 & 0 & 0 & 0 & 0 & 1 \end{bmatrix} \times \epsilon \quad (22)$$

where $\epsilon \gg 0$.
The Kalman prediction is obtained as given below:
$$\hat{x}_{k+1|k} = F\hat{x}_{k|k}, \quad \hat{x}_{0|0} = 0 \quad (23)$$
$$\hat{z}_{k+1|k} = H\hat{x}_{k+1|k} \quad (24)$$

The Kalman correction is obtained as given below:
$$\hat{x}_{k+1|k+1} = \hat{x}_{k+1|k} + K_{k+1}\tilde{z}_{k+1|k} \quad (25)$$
$$\text{Where, } \tilde{z}_{k+1|k} = z_{k+1} - \hat{z}_{k+1|k} \quad (26)$$

In (25), $K_k$ called the Kalman Gain, is computed as

$$K_{k+1} = P_{k+1|k}H^T\left[HP_{k+1|k}H^T + R\right]^{-1} \quad (27)$$

The covariance matrix $P$ is updated as given below:
$$P_{k+1|k} = FP_{k|k}F^T + Q, P_{0|0} = P_0 = E\{x(0)x^T(0)\} \quad (28)$$

In the event of face not being detected by the Haar classifier, the tracker reinitializes itself. Fig. 12 shows the face detection using Haar classifier while Fig. 13 shows the estimated face position using a Kalman filter. Using the Kalman based tracking, the search areas for face get reduced and hence the Haar classifier is able to detect at a faster rate without loss in accuracy as revealed in Table VI.

TABLE III
EFFECT OF KALMAN FILTER ON FACE DETECTION SPEED

|  | Speed (fps) | tpr | fpr |
|---|---|---|---|
| **Without Kalman Filter** | 8.4 | 98.12% | 2.02% |
| **With Kalman Filter** | 9.6 | 97.45% | 2.90% |

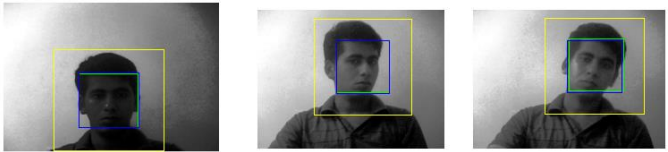
**Fig.12 ROT selection after measurements obtained from Haar Classifier**

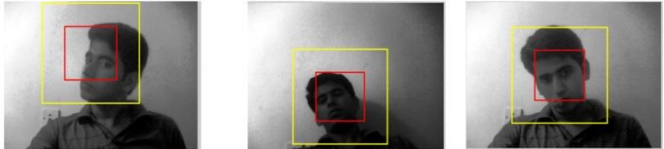
**Fig. 13 Face detection after tracking of ROT using Kalman filter**

## VI. REAL-TIME EYE DETECTION

Once the face is localized and ROI has been remapped, the next step is to detect the eyes in the ROI. Subsequently the detected eye is classified as open or closed to obtain PERCLOS. In this paper the eye detection has been carried out by a PCA based algorithm during day whereas Block LBP features are used for eye detection during night driving.

### A. PCA based eye detection

PCA [58] based eye detection has been used for day driving owing to its superior real-time performance [59]. The ROI is resized to a resolution of 200x70 using bi-cubic interpolation. Block wise search is carried out in each sub-window of size 50x40 pixels with a 10% overlap. The training algorithm is stated briefly in Table IV.

TABLE IV
PCA BASED EYE TRAINING ALGORITHM

1. Obtain eye images $I_1, I_2, I_3 \ldots I_P$ (for training), each of dimension $N \times M$
2. Represent every image $I_i$ of that class as a vector $\Gamma_i$ (of dimension $NM \times 1$)
3. Compute the average eye vector using
$$\psi = \frac{1}{P}\sum_{i=1}^{P}\Gamma_i \quad (29)$$
4. Subtract the mean eye from each image vector $\Gamma_i$
$$\phi_i = \Gamma_i - \psi \quad (30)$$
5. The estimated covariance matrix, $C$ is given by:
$$C = \frac{1}{P}\sum \phi_n \phi_n^T = AA^T \quad (31)$$
Where $A = [\phi_1, \phi_2 \ldots \phi_P]$ ($N^*M \times P$ matrix)
As C is very large, compute $A^TA$ ($P \times P$) instead as $P \ll N$
6. Compute the eigenvectors $v_i$ of $A^TA$.
$$\sigma_i u_i = Av_i \quad (32)$$

Using the equation (32) Eigenvectors $u_i$ of $AA^T$ are obtained
7. Depending on computational capacity available, keep only $K$ eigenvectors corresponding to the $K$ largest Eigen values. These $K$ eigenvectors are the Eigen eyes corresponding to the set of $M$ eye images
8. Normalize these $K$ eigenvectors

The training is carried out using 460 eye images of size 50x40 recorded from several subjects under different illumination levels. The detection is carried out by the method as in Table V.

TABLE V
PCA BASED EYE DETECTION ALGORITHM

The region of interest (ROI), obtained from detected face, is resized to 200x70 matching with the resolution of training images
10% overlapping windows, $\Gamma$ of size 50x40 are used for testing. Then, $\phi$ is obtained from the mean image $\psi$ as
$$\phi = \Gamma - \psi \quad (33)$$
Compute
$$\widehat{\phi} = \sum_{1}^{k} w_i\, u_i \quad (34)$$
The error, $e$ is computed as
$$e = \|\phi - \widehat{\phi_j}\| \quad (35)$$
The window corresponding to the minimum error, $e$ is selected as eye

To evaluate the performance of eye detection, a test was conducted with 300 face images having 200 open and 100 closed eyes taken from the dataset as mentioned earlier.

Fig. 14 shows some Eigen eyes obtained from the training set. Fig. 15 shows some detection results using PCA. From the Table V, it is clear that the PCA technique is quite accurate for detecting the eye from the ROI.

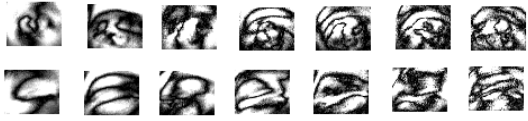
**Fig. 14 Sample Eigen eyes**

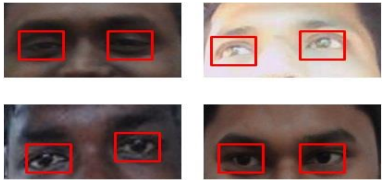
**Fig. 15 Some detection results using PCA**

TABLE VI
EYE DETECTION RESULTS USING PCA

| tp | tn | fp | fn | tpr | fpr |
|---|---|---|---|---|---|
| 198 | 97 | 3 | 2 | 98.5% | 2.02% |

*B. Detection of eye in NIR images*

The PCA based method failed to be robust during night driving conditions, where passive NIR lighting is used to illuminate the face. A low cost Gallium Arsenide NIR illuminator is designed for the purpose. The Block-LBP histogram features is used [60] for detection of eyes in NIR images. Illumination invariance property of LBP features is a major advantage in using them for this purpose.

*C. LBP features*

Due to its discriminative power and computational simplicity, LBP texture operator has become a popular approach in various applications. The LBP operator is defined as a gray-scale invariant texture measure, derived from a general definition of texture in a local neighborhood. It labels the pixels of an image by thresholding the neighborhood of each pixel and considers the result as a binary number. The middle point in a 3x3 neighborhood is set as the threshold, $i_c$ and it is compared with the neighborhood intensity values, $i_n$ for $n = 0, ... ,7$ to obtain the binary LBP code.

$$LBP = \sum_{n=0}^{7} s(i_n - i_c)2^n \quad (36)$$

$$\text{Where,} s(x) = \begin{cases} 1, x \geq 0 \\ 0, x < 0. \end{cases} \quad (37)$$

*D. Block LBP histogram*

For eye detection, a global description of eye region is required. This is achieved using Block LBP histogram. In the LBP approach for texture classification, the occurrences of the LBP codes in an image are collected into a histogram. The classification is then performed by computing simple histogram similarities. However, these result in loss of spatial information. One way to overcome this is the use of LBP texture descriptors to build several local descriptions of the eye and combine them into a global description. These local feature based methods are more robust against variations in pose or illumination than holistic methods. The figure below shows how an LBP feature is computed.

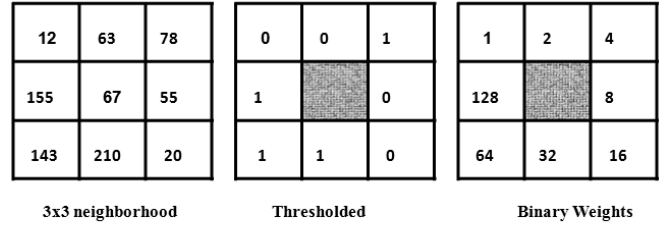
**Fig. 16 LBP Calculation**

The basic methodology for LBP based face description proposed by Ahonen *et al.* [60] is as follows:
- The facial image is divided into local regions and LBP texture descriptors are extracted from each region independently.
- The descriptors are then concatenated to form a global description of the face, as shown in Fig. 17.

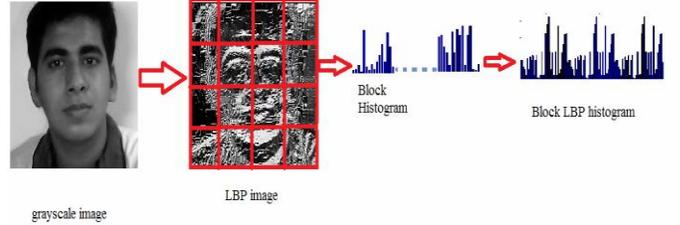
**Fig. 17 Block LBP histogram feature calculation**

*E. NIR face database creation*

For robust training of LBP features, it is required to create a proper dataset under NIR illumination. A dataset of facial images were created under laboratory conditions with different facial orientation and eyelid positions.

*F. Eye Detection Algorithm using block LBP Features*

The use of local histograms and pixel level values of LBP to get the feature vector is given in Table VII. In the detection phase, LBP histogram features are obtained in the localized eye region and then projected into the PCA feature space to get the energy components.

TABLE VII
ALGORITHM: BLOCK LBP

Training Phase
- Eye images are resized to 50x40 resolution and each image is divided into sub-blocks of size 5x4
- The LBP feature values of the sub-blocks are found and 16 bin histogram of each block is calculated
- The histogram of each block is arranged to form a global feature descriptor
- PCA is carried out on the feature vectors and 40 Eigen vectors having largest Eigen values were selected.

Detection Phase
- ROI from face detection stage is obtained
- For each sub window the Block- LBP histogram is found and is projected to Eigen space

- The sub window with minimum reconstruction error is found
- If the reconstruction error is less than a threshold it is considered as a positive detection

300 prerecorded NIR images of the eye are used for training. The algorithm is tested on the whole database. Fig. 18 shows some eye detection results using block LBP features.

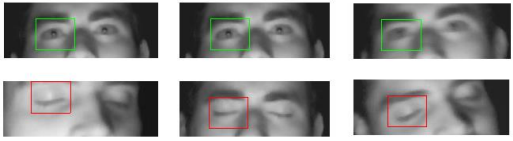

**Fig. 18 Eye localization using LBP**

The results as in Table VIII reveal its robustness over PCA for NIR images. However, as evident from Table IX, PCA performs faster than block LBP feature based method, and has been used for eye detection during day driving.

TABLE VIII
RUN-TIME COMPARISON OF PCA AND BLOCK LBP

| Algorithm | Speed (fps) |
|---|---|
| PCA | 7.8 |
| LBP | 6.6 |

TABLE IX
COMPARISON OF PCA AND BLOCK LBP FOR NIR IMAGES

|  | tpr | fpr |
|---|---|---|
| PCA | 80% | 10% |
| LBP | 98% | 2% |

## VII. EYE STATE CLASSIFICATION

For accurate estimation of PERCLOS, the detected eye needs to be accurately classified into open or closed state. SVM is reported to be a robust binary classifier [61] and is used here for the purpose.

### A. SVM Training

SVM [62] is a learning method applied for binary data classification problems. In the present case, the eye state classification is a binary classification problem where the class labels include open eye class and closed eye class. The weights obtained from the eye detection part, along with the ground truth, are used for the training input of the SVM. The energy components are fed into the SVM which returns the class number.

### B. Results

The training of SVM has been carried out with 460 images taken from the database of 8 subjects created using normal and NIR illumination. Some training images are shown in Fig. 19. The testing is carried out using another 1700 images taken from the same database. The block LBP histogram feature transform is carried out on the training images. Then PCA is carried out on the LBP feature vectors. Now the weight vectors corresponding to each sample is found by projecting the samples to PCA subspace. These weights along with the ground truth are used for SVM training. The training is carried out with different kernels and accuracy levels and the classification results are shown in Table X.

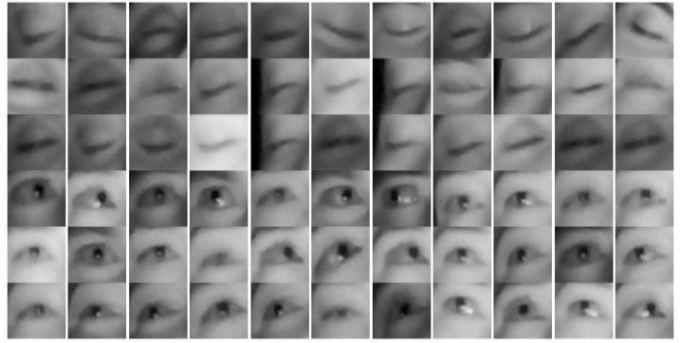

**Fig. 19 Training images for eye state classification using SVM**

TABLE X
DETECTION RESULTS WITH SVM

| Kernel function | tp | tn | fp | fn | tpr | fpr |
|---|---|---|---|---|---|---|
| Linear SVM | 838 | 808 | 42 | 12 | 98.58% | 4.94% |
| Quadratic | 827 | 765 | 85 | 23 | 97.29% | 10% |
| Polynomial | 848 | 807 | 43 | 2 | 99.76% | 5.32% |

The results reveal that the performance of the SVM is consistent under normal and NIR illuminated images. The real-time algorithm is coded into an SBC having Intel Atom processor, with 1.6 GHz processing speed and 1 GB RAM. The overall processing speed is found to be 9.5 fps, which is sufficient for estimating PERCLOS accurately.

## VIII. TESTING

### A. System Description

The alertness monitoring system consists of a single USB camera having maximum frame rate of 30 fps at a resolution of 640×480 pixels. The camera is placed directly on the steering column just behind the steering wheel to obtain the best view of the driver's face. The SBC is powered at 12 V DC from the car supply. The typical current drawn from the input source is approximately 1200 mA. The approximate typical power drawn from the supply is 15 W. A voltage regulator unit, comprising of IC LM317 along with some resistors, a capacitor and an inductor, is used before the input to remove high voltage spikes from the car supply. An NIR lighting arrangement, consisting of a matrix of 3×8 Gallium Arsenide LEDs, is also connected across the same supply in parallel with the Embedded Platform. The NIR module is operated at 10 V DC and draws a typical current of 250 mA. The lighting system is connected through a Light Dependent Resistor (LDR), to automatically switch on the NIR module in the absence of sufficient illumination. A seven inch LED touch screen is used to display the results. Fig. 20 shows the different modules used in the system. The speed of the algorithm was found to be 9.5 fps, which is very acceptable for estimation of PERCLOS.

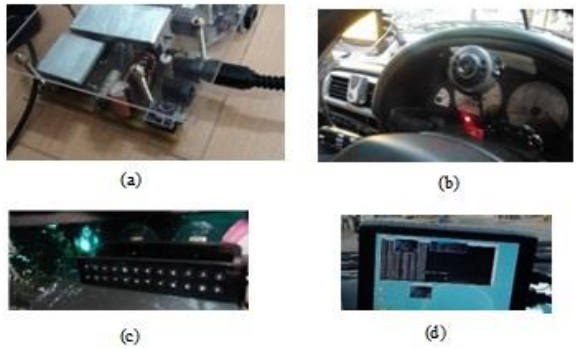

**Fig. 20 (a) surge protector circuit (b) camera on dashboard (c) NIR lighting system (d) LCD screen on dashboard**

*B. Testing under Simulated Driving*

The algorithm is tested systematically with simulated driving in the laboratory as well as with on-board driving. Twenty male subjects within the age group of 25-40 years age participated in the test in the actual as well as simulated driving. A low cost driving simulator was used for the purpose of evaluation of the system before actual deployment and field testing. Fig. 21 shows the driving simulator. The subject under test was asked to drive on the simulator and different road conditions were simulated on a big screen which was kept directly in front of the driver.

Fig. 22 shows some images of testing under simulated driving condition. The testing in the driving simulator showed accurate results with about 2% error in PERCLOS estimation. However, the on-board conditions are quite different from the laboratory conditions and therefore testing in practical driving scenarios was carried out. Twenty on-board tests were performed separately for day and night to test the robustness of our algorithm in actual driving conditions.

*C. Testing under Day-driving Conditions*

The main objective of the testing was to check the performance of the system at different lighting conditions during day time. The testing was carried out on a rough followed by a smooth road (highway). The face and eye detection was found to be accurate for both smooth as well as jerky road conditions. However, during the midday, when the light fell directly on the camera lens, the camera sensor got saturated and image features were lost as shown in Fig. 24. The problem was rectified by using a camera with a superior image sensor to remove the saturation effect. Fig. 25 shows a sample detection result during the start of the test. The analysis of the accuracy of the system has been carried out offline from the on-board recorded videos to evaluate the performance of the system. The ground truth was manually marked. The face detection rate, when the driver was looking straight, is found to be 98.5% whereas the eye detection rate was 97.5% on an average. With head rotations of the driver, the face detection rate dropped down to 95% whereas that for eyes was 94%. The eye state classification was found to be 97% accurate. The overall false alarming of the algorithm was found to be around 5%.

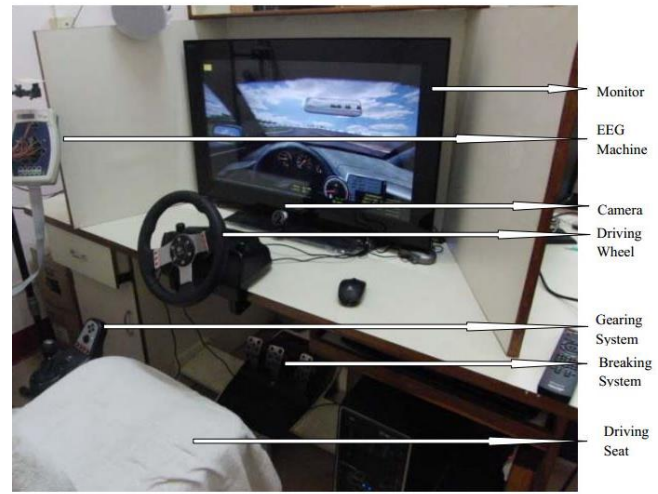

**Fig. 21 Driving simulator**

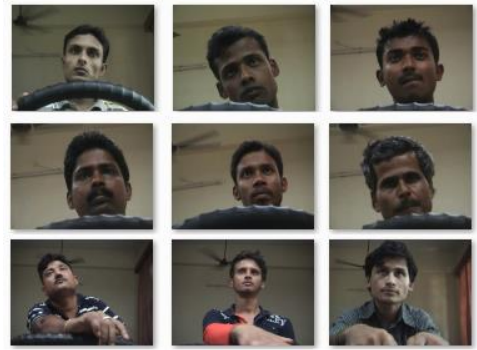

**Fig. 22 Testing under simulated driving with different subjects**

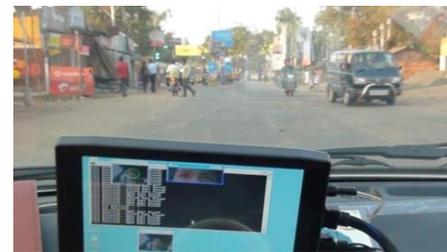

**Fig. 23 On-board testing under day driving conditions**

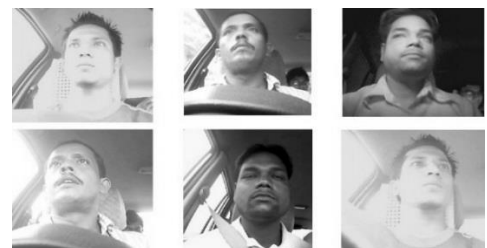

**Fig. 24 On Board testing with different subjects**

The error in PERCLOS estimation was calculated by the difference of true PERCLOS computed manually with the PERCLOS value obtained from the algorithm. The error is found out to be 4.5% approximately. The PERCLOS values are shown in the display for every minute, on a running average window basis of 3 minutes. The main issues faced during the day driving test are the image saturation due to sunlight directly falling on the camera lens and noise due to vehicle vibration. There is still scope of research left in addressing the issues.

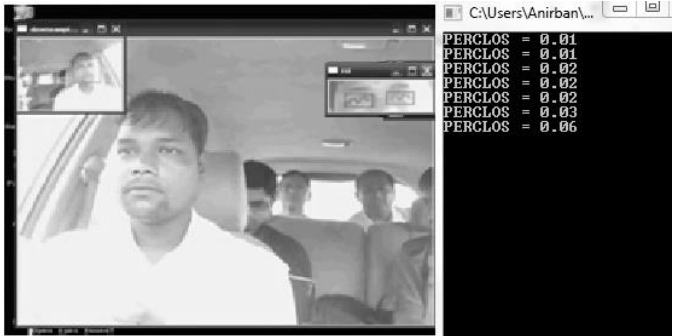
Fig. 25 Sample detection on-board during daytime

### D. Testing under Night Driving Conditions

As per literature, most of the accidents due to drowsiness of the driver occur during the night [63] [64]. Hence, separate on-board testing during night was carried out. The NIR lighting system is automatically switched on by the LDR. The illumination level was enough to detect faces properly providing a clear NIR image without distracting the driver. The tests were performed on a highway. The objectives of the test was to check the performance of the NIR lighting system and to find out the performance of the algorithm in detecting face and eyes during night driving condition. Except for a few cases of malfunctioning owing to the lights coming from the other vehicles and the streetlights, the system was found to be robust in detecting the face and eyes. The face detection rate, when the driver was looking straight, is found to be 98.2% whereas the eye detection rate was 97.1%. With head rotations of the driver, the face detection rate dropped down to 94.5% whereas that for eyes was 94.2%. The eye state classification was found to be 98% accurate. The overall false alarming of the algorithm was found to be around 5.5%. The error in PERCLOS estimation was calculated by the difference of true PERCLOS computed manually with the PERCLOS value obtained from the algorithm. The error is found out to be 4.9% approximately. Fig. 26 shows a detection during night driving. The NIR lighting system works well for the test duration. From the above on-board tests, it is evident that this system can be used as a safety precaution system which might prevent a number of road accidents due to drowsiness.

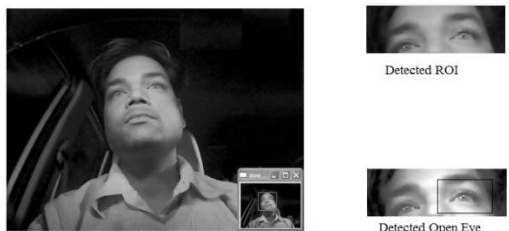
Fig. 26 Detection of eyes in NIR lighting

### IX. CONCLUSION

In this paper, a robust real-time system for monitoring the loss of attention in automotive drivers has been presented. In this approach, the face of the driver is detected at a lower resolution using a Haar classifier. An optimal down Sampling Factor (SF) of 6 is chosen as a trade-off between speed and accuracy. The in-plane and off-plane rotations of the driver's face have been compensated using affine and perspective transformations of the input frame respectively. To compensate the effect of variation in illumination, BHE has been performed. Further, a superscribed rectangular region over the detected face has been tracked using a Kalman Filter thereby making use of the temporal information resulting in reduction of the search space and improvement in real-time performance. An ROI based on face morphology has been marked and remapped onto the original frame where PCA is used during day and LBP feature is used during night with NIR illumination to localize the eyes. Finally SVM is used to classify the eye states into open and closed to compute the PERCLOS values over a window of 3 minutes interval. Linear SVM shows a hit rate of 98.6% while quadratic SVM has a hit rate of 97.3%. The overall speed of the algorithm is found to be 9.5 fps, which is good enough for correctly estimating the state of eye. The algorithm has been cross-validated using EEG signals. On-board as well as testing has been carried out for both day as well as night driving. The system was found to be quite robust both in terms of speed and accuracy. There is further scope of research in the detection of eyes occluded by spectacles.

### X. ACKNOWLEDGMENT

The funds received from the Department of Electronics and Information Technology, Government of India for this research work is gratefully acknowledged. The authors would like to thank the subjects for voluntary participation in the experiment for creating the database.

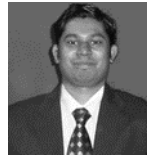

**Anirban Dasgupta** has received his B.Tech. (Hons.) degree from National Institute of Technology (NIT), Rourkela, India in 2010 and now pursuing M.S. degree from Indian Institute of Technology (IIT) Kharagpur, India. His current research interests include computer vision, embedded system and image processing.

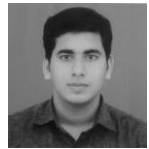

**Anjith George** has received his B.Tech. (Hons.) degree from Calicut University, India in 2010. and M-Tech degree from Indian Institute of Technology (IIT) Kharagpur, India in 2012. Presently he is pursuing Ph.D. from IIT Kharagpur. Kharagpur. His current research interests include real time computer vision and its applications.

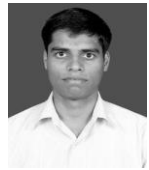

**S L Happy** has received the B.Tech. (Hons.) degree from Institute of Technical Education and Research (ITER), India in 2011. Now he is pursuing the M. S. degree from Indian Institute of Technology (IIT) Kharagpur, India. His research interests include pattern recognition, computer vision and facial expression analysis.

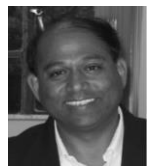

**Aurobinda Routray** is a professor in the Department of Electrical Engineering, Indian Institute of Technology, Kharagpur. His research interest includes non-linear and statistical signal processing, signal based fault detection and diagnosis, real time and embedded signal processing, numerical linear algebra, and data driven diagnostics.